\documentclass[conference]{IEEEtran}
\IEEEoverridecommandlockouts

\usepackage[inline]{enumitem}
\usepackage{algorithmic}
\usepackage{graphicx}
\usepackage{textcomp}
\usepackage{xcolor}
\usepackage{witharrows}
\usepackage[flushleft]{threeparttable}
\usepackage{tablefootnote}
\usepackage{array}
\newcolumntype{P}[1]{>{\centering\arraybackslash}p{#1}}
\usepackage{multicol}
\usepackage{multirow}
\usepackage{tabulary}
\usepackage{colortbl}
\definecolor{mygray}{gray}{0.90}
\usepackage{makecell}
\usepackage{booktabs}

\usepackage{subcaption}

\usepackage{amsmath}
\usepackage{amssymb} 

\usepackage{pifont}
\newcommand{\xmark}{\raisebox{0.2ex}{\scalebox{0.85}{\ding{55}}}}

\usepackage{stmaryrd}
\usepackage{float}
\usepackage{arydshln}
\colorlet{mygray}{gray!15!white}
\usepackage[switch,columnwise]{lineno}

\usepackage{cite}

\usepackage{url,hyperref,microtype}
\hypersetup{
    colorlinks=true,
    citecolor=blue,
    linkcolor=blue,
    filecolor=magenta,      
    urlcolor=black,
}

\def\BibTeX{{\rm B\kern-.05em{\sc i\kern-.025em b}\kern-.08em
    T\kern-.1667em\lower.7ex\hbox{E}\kern-.125emX}}

\begin{document}

\title{ReFace: Reorganizing Facial Spatiotemporal Representations for Improved Pain Assessment
}

\author{\IEEEauthorblockN{Stefanos Gkikas}
\IEEEauthorblockA{\textit{Honda Research Institute Japan} \\
Wako City, Japan \\
stefanos.gkikas@jp.honda-ri.com}
\and
\IEEEauthorblockN{ Yu Fang}
\IEEEauthorblockA{\textit{Honda Research Institute Japan} \\
Wako City, Japan  \\
yu.fang@jp.honda-ri.com}
\and
\IEEEauthorblockN{Christian Arzate Cruz}
\IEEEauthorblockA{\textit{Honda Research Institute Japan} \\
Wako City, Japan  \\
christian.arzate@jp.honda-ri.com}
\and

\IEEEauthorblockN{Muhammad Umar Khan}
\IEEEauthorblockA{\textit{BioSIS (Biosensing \& Intelligent Systems) Lab}\\ 
\textit{Centre for Intelligent Computing and Systems} \\
\textit{University of Canberra}\\
Canberra, Australia\\
umar.khan@canberra.edu.au}

\and
\IEEEauthorblockN{Raul Fernandez Rojas}
\IEEEauthorblockA{\textit{BioSIS (Biosensing \& Intelligent Systems) Lab} \\
\textit{Centre for Intelligent Computing and Systems} \\
\textit{University of Canberra} \\
Canberra, Australia \\
raul.fernandezrojas@canberra.edu.au}

}

\maketitle

\begin{abstract}

Automatic pain assessment from facial video remains challenging due to the spatial heterogeneity of pain-related facial cues. This study proposes \textit{ReFace}, a spatial reorganization pipeline that divides facial input into four spatial quadrants before tokenization, rather than processing the entire face as a single region. Evaluated on the \textit{AI4Pain} dataset, the proposed approach achieves $56.00\%$ accuracy on the test set using video only, achieving the highest reported accuracy under the fixed AI4Pain benchmark protocol among the compared methods. Notably, the four-quadrant configuration processes the same total pixel budget as the full-face input, yet achieves higher accuracy, suggesting that spatial reorganization can improve performance under the proposed tokenization design. A single quadrant region, processing just one quarter of those pixels, remains competitive at a fraction of the computational cost.

\end{abstract}


%\linenumbers
\section{Introduction}

Although pain is one of the most prevalent health conditions in contemporary healthcare, assessment of it remains mostly based on patients' own reports. Although many types of tools exist for assessing pain (numeric rating scales, verbal descriptors, etc.) that have evolved very little since the last few decades; they are all subjective and therefore provide an inherent limitation in terms of the clinical relevance of using such measures as well as the risk of undertreating pain in those who cannot report it and the risks of overprescribing opioids in those that can report it. They also limit our ability to assess the objective success or failure of treatment in real-time \cite{gkikas_tsiknakis_slr_2023, gkikas_phd_thesis_2025}.
Pain is the leading cause of years lived with disability worldwide according to the Global Burden of Disease study \cite{james_abate_2018}, with back pain, musculoskeletal disorders, and neck pain among the primary contributors \cite{usa_bdc_2013}. In the United States, the annual costs associated with pain exceed those of heart disease, cancer, and diabetes combined \cite{gaskin_richard_2012}, while in Europe chronic pain accounts for $3\%$ to $10\%$ of gross domestic product \cite{breivik_eisenberg_2013}. Pain also directly impairs attention, functional performance, and cognitive processing speed \cite{khera_rangasamy_2021}, and is strongly associated with opioid overuse, addiction, and psychological disorders \cite{dinakar_stillman_2016}. Opioid analgesics remain the most frequently prescribed treatment \cite{kaye_jones_2017}, despite well-documented risks of addiction and overdose \cite{stampas_pedroza_2020}, as well as side effects including sedation, depression, and anxiety that substantially reduce quality of life \cite{benyamin_trescot_2008}.

Pain evaluation becomes significantly harder when people cannot report their own pain. This includes critically sick adults; infants; elderly adults living in residential care home settings; and adult patients who are cognitively impaired or have communication disorders \cite{puntilo_staannard_2022, abdulla_adams_2013, meehan_mcrae_1995}. Clinicians will use observational measures (\textit{e.g.,} how someone behaves) and physiological signals (\textit{e.g.,} heart rate and skin conductance) \cite{rojas_brown_2023, gkikas_tsiknakis_slr_2023}, rather than relying solely on what the patient reports about their pain level. While this provides valuable information, there is still subjectivity in the clinician's interpretation of the same signs, depending on how he/she evaluates them. Therefore, developing an automated system to recognize pain would be beneficial for continuous, objective observation \cite{gkikas_tsiknakis_slr_2023}.

Video facial analysis is best suited for automatically assessing a person's pain level due to its non-invasive, contactless nature and its ability to capture behavioral cues of pain through facial expressions. The above-mentioned behaviors have been demonstrated throughout various patient groups \cite{rojas_brown_2023}. Behavioral cues of pain include, but are not limited to, lowering the brow, closing the eyes, wrinkling the nose, and parting the lips. Since facial expressions are widely used as behavioral cues of pain, video-based methods account for the majority of research related to the automation of pain assessment. Previous early models were developed using manually created facial action units \cite{werner_hamadi_2014, werner_hamadi_walter_2017}. Following this, CNNs were used for temporal modeling \cite{huang_dong_2022}. More recent models have employed transformer architecture \cite{gkikas_tsiknakis_embc, nguyen_yang_2024, gkikas_rojas_painformer_2025}. 
Methods that incorporate both video and physiological signals have demonstrated improved performance, including approaches based on facial video with heart-rate signals \cite{gkikas_tachos_2024}, video with fNIRS \cite{gkikas_tsiknakis_painvit_2024, gkikas_arzate_pain_icmi_2026}, and EDA--fNIRS or EDA--ECG fusion \cite{khan_chetty_2026,farmani_bargshady_2025}.
However, many of these multimodal approaches also require additional sensors or infrastructure for deployment in real-world settings. A shortcoming of all previously described methods is that the entire face is treated as a single spatial area; therefore, the model must identify where relevant features exist without explicit guidance. Broader overviews of the field are provided in \cite{gkikas_tsiknakis_slr_2023, khan_umar_2025}.

\textit{ReFace} addresses this by reorganizing the facial input into four spatial quadrants before tokenization, combining their spatiotemporal information into a unified representation. Experiments on the \textit{AI4Pain} dataset show that this reorganization improves recognition over an equivalent full-face baseline while using the same number of pixels. Notably, a single quadrant, at one-quarter of the pixel budget, also achieves competitive performance at substantially lower cost.

\section{Related Work}
\label{related_work}
In recent years, many efforts have been made to advance the field of automatic pain assessment using facial videos. The most significant of them was the transition from handcrafted features to the development of complex deep neural networks. Earlier methods were based on measuring distances between facial landmarks, estimating head pose, and tracking motion (optical flow) to quantify facial movement \cite{werner_hamadi_2014, werner_hamadi_walter_2017}. More recently, convolutional neural networks (CNNs) have demonstrated that training a model directly on raw video yields a representation that is richer in information and more robust. Researchers have achieved strong performance across a variety of architectures, including 3D CNNs trained for varying durations \cite{tavakolian_hadid_2019} and CNNs incorporating self-attention \cite{huang_dong_2022}. In addition, researchers used transformer-based models to improve upon the previous state-of-the-art architecture. This resulted in good performance when training hybrid CNN--Transformer architectures and vision transformers on raw video sequences \cite{gkikas_tsiknakis_embc, nguyen_yang_2024}.
Other studies have explored pain assessment using ECG and demographic factors \cite{gkikas_chatzaki_2022,gkikas_chatzaki_2023}, synthetic thermal/RGB videos \cite{gkikas_tsiknakis_thermal_2024}, and compact biosignal representations \cite{gkikas_tiny_2025}.

Only a few researchers have investigated how specific facial regions contribute to an individual's expression of pain. Huang \textit{et al.} \cite{huang_xia_li_2019} developed a multi-stream CNN that segmented facial regions and assigned learned attention weights to each region, reflecting their relative contributions to pain expression. The authors in \cite{gkikas_tachos_2024} incorporated multiple facial regions into a model, developing a transformer-based multimodal approach that tiles the input frame into multiple regions, along with the full frame, and processes each region and the full frame through a hierarchical inner--outer transformer before fusing the resulting embeddings. Although that method encodes each facial region separately and then fuses the representations, the present work reorders all facial regions into a single unified spatiotemporal tensor before tokenizing it, embedding spatial relationships in the input data itself rather than recovering them after each stream has encoded its respective region.

The \textit{AI4Pain} benchmark \cite{ai4pain_2024, rojas_hirachan_2023} has recently become a common platform for evaluating pain recognition algorithms. Studies conducted on this dataset range from handcrafted SVM-based systems \cite{ai4pain_2024} to deep transformer-based frameworks that can operate on video alone, on functional near-infrared spectroscopy (fNIRS), or on both \cite{gkikas_tsiknakis_painvit_2024, nguyen_yang_2024, gkikas_rojas_painformer_2025}. Among these, one video-only study \cite{nguyen_yang_2024} reported $55.00\%$ accuracy on the test set, while a multimodal framework \cite{gkikas_rojas_painformer_2025} achieved $55.69\%$ accuracy by combining video and functional near-infrared spectroscopy (fNIRS). 
Recent physiological-signal studies have also investigated EDA-based visual representations \cite{gkikas_kyprakis_eda_2025}, respiration-based pain recognition \cite{gkikas_kyprakis_resp_2025}, and fNIRS hemoglobin-difference modeling \cite{bargshady_aziz_2025, gkikas_arzate_eeite_pain_2026}. Related modality-agnostic multimodal frameworks have also been explored for cognitive workload assessment \cite{gkikas_arzate_workload_acii_2026}.

%%%%%%%%%%%%%%%%%%%%%%%%%%%%%%%%%%%%%%%%%%%%%%%%%%%%%%%%%%%%%%%%
\section{Methodology}
This section presents the proposed face reorganization pipeline and tokenization framework, along with the preprocessing and architecture used for pain assessment.

\subsection{Pre-processing}
The pre-processing stage includes face detection in video frames using the MTCNN detector \cite{zhang_2016}, which localizes the facial region through a cascade of convolutional neural networks. For each frame, the detected facial bounding box was cropped and resized to $224\times224$ pixels before any spatial partitioning was applied. 
The same face detection and resizing procedure was used for all configurations, ensuring that the full-face, half-face, single-quadrant, and four-quadrant experiments remain controlled with respect to pre-processing.

\subsection{Reorganization}
\label{sec:reorganization}
Facial areas make different contributions in the process of pain expression; each facial area (the brow, eyes, nose, and mouth) represents a separate aspect of the same overall behavioral process; when the whole face is considered as one single ``token'' for processing all of the region-specific information, it is treated equally. 
The proposed method challenges current facial analysis approaches by introducing a reorganization pipeline that explicitly separates the input into spatially defined regions prior to tokenization. The information fusion across facial regions occurs at the representation level.

\subsubsection{Image Segmentation}
Given a facial video of $T_{\max}=300$ frames ($10s \times 30$ FPS) at spatial resolution $224\times224$ pixels, the face is partitioned into $K=4$ non-overlapping quadrant regions: top-left (\textit{TL}), top-right (\textit{TR}),
bottom-left (\textit{BL}), and bottom-right (\textit{BR}).
Each quadrant is obtained by partitioning the image along both spatial dimensions, resulting in regions with a resolution of $112\times112$ pixels. The four regions tile the full face without overlap or gaps:
\begin{equation}
\begin{split}
   F &\in \mathbb{R}^{T_{\max} \times 224 \times 224 \times 3}
   \;\longrightarrow\;
   \{F^{(k)}\}_{k=1}^{K}, \\
   F^{(k)} &\in \mathbb{R}^{T_{\max} \times 112 \times 112 \times 3}.
\end{split}
\end{equation}

\subsubsection{Token Mixing}
The temporal and spatial information from all $K$ regions is fused into a single unified tensor through five sequential steps.

\paragraph{Frame Sampling}
Each region $k$ is temporally subsampled at a uniform stride $s$, retaining
$T = \lfloor T_{\max} / s \rfloor$ frames. The stride $s$ varies across
experiments to study the effect of temporal resolution on recognition
performance and computational cost, while all other components remain fixed.

\paragraph{Axis Folding}
The $T$ sampled frames of each region are stacked along the channel dimension,
converting the temporal axis into additional channels:
\begin{equation}
   F^{(k)} \in \mathbb{R}^{T \times 112 \times 112 \times 3}
   \;\longrightarrow\;
   \hat{F}^{(k)} \in \mathbb{R}^{112 \times 112 \times 3T}.
\end{equation}
This folding preserves spatial structure while collapsing the temporal axis,
so that the tokenizer operates on a standard 2D spatial grid regardless of temporal length.

\paragraph{Channel Concatenation}
The $K$ folded region tensors are concatenated along the channel dimension to
produce the unified input tensor:
\begin{equation}
   X = \mathrm{Cat}_{C}\!\big(\hat{F}^{(1)},\ldots,\hat{F}^{(K)}\big)
   \in \mathbb{R}^{112 \times 112 \times C},
   \quad C = 3TK,
\end{equation}
where each region contributes $3T$ channels.

\paragraph{Flatten Spatial Axes}
The spatial axes of $X$ are flattened into a sequence of $N = H \times W$
tokens, converting the 2D spatial grid into a 1D sequence suitable for
attention-based processing:
\begin{equation}
   X \in \mathbb{R}^{H \times W \times C}
   \;\longrightarrow\;
   X \in \mathbb{R}^{N \times C}, \qquad N = H \times W.
\end{equation}

\paragraph{Fourier Features}
Geometric information is incorporated by encoding the normalized spatial
coordinates $p \in [-1,1]^2$ of each token using Fourier features with $K_f$
frequency bands and maximum frequency $f_{\max}$:
\begin{multline}
\gamma(p) = \big[
\sin(\pi s_1 p),\ \cos(\pi s_1 p),\ \ldots,\ \\
\sin(\pi s_{K_f} p),\ \cos(\pi s_{K_f} p),\ p
\big],
\end{multline}
where $\{s_k\}_{k=1}^{K_f}$ span $[1, f_{\max}/2]$. For each spatial axis, the encoding contains $K_f$ sine terms, $K_f$ cosine terms, and the raw coordinate, giving $2K_f+1$ features. Since each token has two spatial coordinates, the total positional dimension is $2(2K_f+1)$. Quadrant identity is not encoded as a separate embedding; instead, each quadrant occupies a fixed channel group in the concatenated tensor. Data channels and positional features are concatenated per spatial location to create the token matrix:
\begin{equation}
\mathcal{T} \in \mathbb{R}^{B \times N \times C'}, \qquad
C' = C + 2(2K_f+1),
\end{equation}
where $B$ is the batch size. The resulting token matrix $\mathcal{T}$ is passed to the model presented in the next section, which performs token segmentation, latent processing, and final pooling to obtain the final classification.

\subsection{Architecture}
\label{sec:architecture}
The architectural design is based on a token matrix $\mathcal{T}$ produced by the reorganization pipeline. This architectural design uses an iterative process to encode each input into a compact learned latent space, process a token sequence with cross-attention, and then apply self-attention and feed-forward transformations to the cross-attentive output. In this case, the dimensionality of the learned latent states $d$ will always be the same for all configurations of input; however, the token dimension $C' = C + 2(2K_f+1)$, where $C = 3TK$ ($C$ being dependent on $TK$ and therefore the number of tokens in the token matrix), can change depending on how many regions $K$ are used when processing each temporal unit at time $t$, and/or depending on the stride size $s$. Because the number of parameters in the cross-attention layers and their computational cost depend on both $K$ and $s$, these two hyperparameters allow control over the computational resources used during training.
This property is reflected in the results tables, where different stride and region configurations report different numbers of parameters and GFLOPs.

\subsubsection{Attention Mechanism}
Multi-head attention (MHA) projects queries, keys, and values into $H_a$ independent heads, computes scaled dot-product attention per head, and concatenates the results:
\begin{equation}
\mathrm{MHA}(Q, K_a, V)
=
\mathrm{Concat}(\mathrm{head}_1,\ldots,\mathrm{head}_{H_a}) W^O,
\end{equation}
\begin{equation}
\mathrm{head}_h
=
\mathrm{softmax}\!\left(\frac{Q_h K_{a,h}^\top}{\sqrt{d_h}}\right) V_h,
\end{equation}
where $K_a$ denotes the key matrix (subscript $a$ distinguishes it from $K$, the number of regions).
Layer normalization precedes each attention and feedforward sublayer, and residual connections add the sublayer input to its output. Cross- and self-attention share the same operator and differ only in whether the context is the token matrix $\mathcal{T}$ or the latent state matrix itself.

\subsubsection{Segment-Latent Processing and Pooling}
\begin{figure*}
\begin{center}
\includegraphics[scale=0.45]{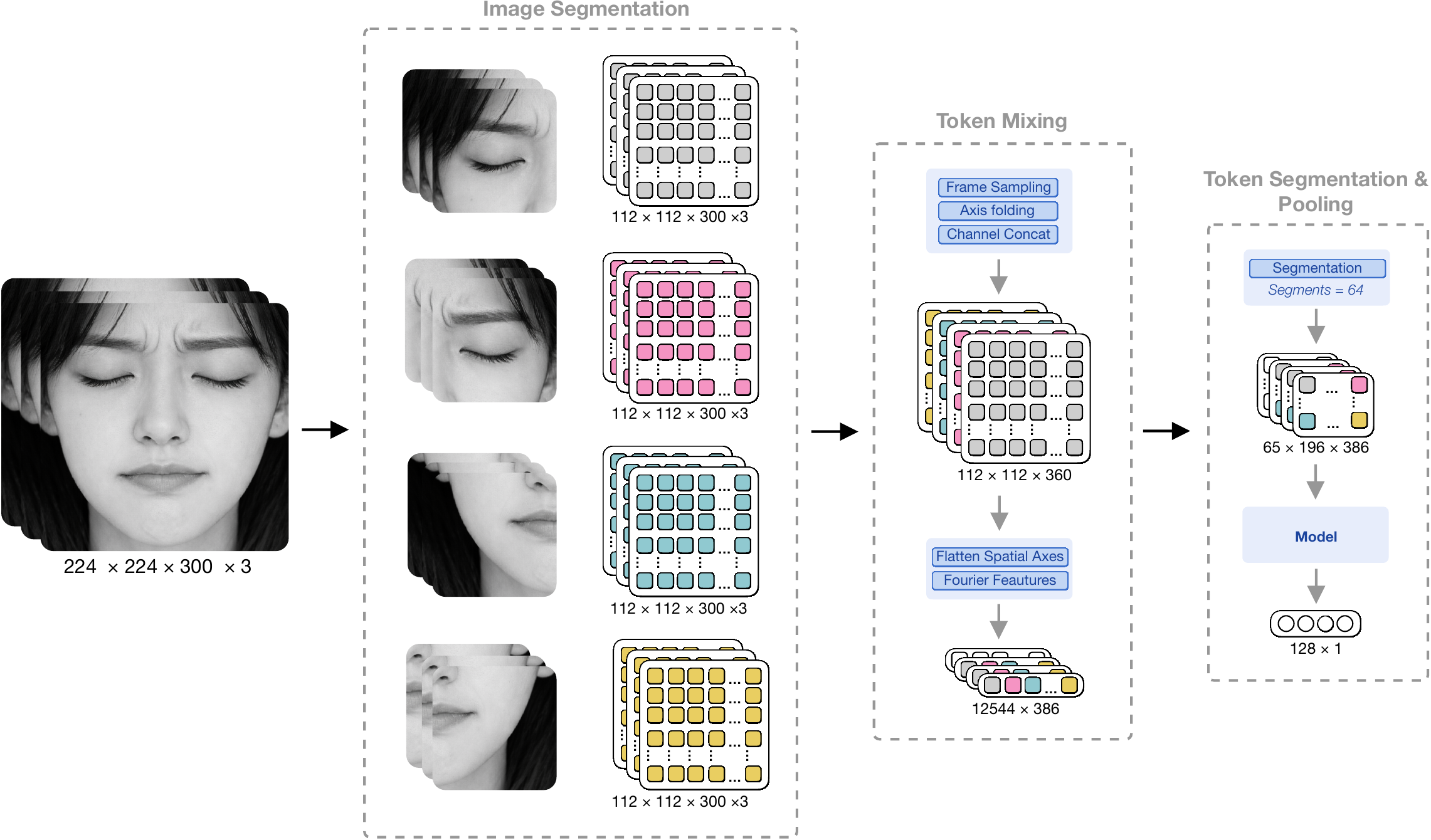} 
\end{center}
\caption{Overview of the proposed reorganization pipeline. The input facial video is spatially partitioned into four quadrant regions via image segmentation. Token mixing fuses regions via frame sampling, axis folding, and channel concatenation into a unified tensor, which is then tokenized using Fourier positional features. Token segmentation partitions the resulting sequence into $S=64$ segments, each aggregated by a dedicated latent state, followed by mean pooling to produce the final representation.}
\label{overview}
\end{figure*}

The token sequence is partitioned into $S=64$ contiguous segments of length
$n_s = \lceil N / S \rceil$, padded to $\tilde{N}=Sn_s$ and reshaped as:
\begin{equation}
\tilde{\mathcal{T}} \in \mathbb{R}^{B \times S \times n_s \times C'}.
\end{equation}
Each segment $s$ is associated with a single latent state
$e^{(0)}_s \in \mathbb{R}^{B \times d}$, initialized from the mean of $M$
learnable parameter vectors $\{\ell_m\}_{m=1}^{M}$:
\begin{equation}
\ell_{\mathrm{init}} = \frac{1}{M}\sum_{m=1}^{M} \ell_m \in \mathbb{R}^{d}.
\end{equation}
Segment-specific representations emerge through segment-local cross-attention,
where each latent state attends exclusively to the tokens of its own segment:
\begin{equation}
e^{(\ell)}_s = e^{(\ell-1)}_s
+ \mathrm{Attn}\!\big(e^{(\ell-1)}_s,\ \tilde{\mathcal{T}}_s\big).
\end{equation}
Stacking all segment states gives
$E^{(\ell)} \in \mathbb{R}^{B \times S \times d}$.
Global information exchange across segments is then performed via self-attention
over this matrix:
\begin{equation}
E^{(\ell)} \leftarrow E^{(\ell)}
+ \mathrm{Attn}\!\big(E^{(\ell)},\ E^{(\ell)}\big),
\end{equation}
applied $R$ times per layer. Segment-local cross-attention is performed efficiently in parallel by temporarily packing all $S$ segments into the batch dimension. After cross-attention, the original batch and segment structure is restored before global self-attention and final mean pooling.
After $L$ layers, the final segment-state matrix
$E^{(L)} \in \mathbb{R}^{B \times S \times d}$ is mean-pooled over segments
and passed through a linear classification head to produce the final
prediction:
\begin{equation}
   \hat{y} = W\,\frac{1}{S}\sum_{s=1}^{S} e^{(L)}_s + b \;\in\; \mathbb{R}^{N_c},
\end{equation}
where $N_c$ is the number of target classes. The model hyperparameters used across all experiments are summarized in
Table~\ref{tab:architecture_details}, and the full pipeline is illustrated
in Figure~\ref{overview}.

%%%%%%%%%%%%%%%%%%%%%%%%%%%%%%%%%%%%%%%%%%%%%%%%%%%%%%%%%%%%%%%%

\begin{table}
\caption{Architectural hyperparameters.}
\label{tab:architecture_details}
\centering
\begin{tabular}{lc}
\toprule
Hyperparameter & Value \\
\midrule
\midrule
Depth ($L$)                           & 1   \\\hdashline
Number of latents ($M$)               & 64  \\\hdashline
Latent dimension ($d$)                & 128 \\\hdashline
Cross-attention heads                 & 1   \\\hdashline
Latent self-attention heads           & 4   \\\hdashline
Cross-attention head dimension        & 64  \\\hdashline
Latent attention head dimension       & 128 \\\hdashline
Self-attention blocks per cross ($R$) & 1   \\\hdashline
Segments ($S$)                        & 64  \\\hdashline
Fourier frequency bands ($K_f$)       & 6   \\\hdashline
Maximum frequency ($f_{\max}$)        & 10  \\
\bottomrule
\end{tabular}
\end{table}

%%%%%%%%%%%%%%%%%%%%%%%%%%%%%%%%%%%%%%%%%%%%%%%%%%%%%%%%%%%%%%%

%%%%%%%%%%%%%%%%%%%%%%%%%%%%%%%%%%%%%%%%%%
\subsection{Augmentation \& Regularization}
Several augmentation and regularization techniques were applied in all experiments. Video frames were augmented using \textit{TrivialAugment} \cite{trivialAugment}, \textit{AugMix} \cite{augmix}, additive noise, and spatial \textit{Masking}. For each video sample, the same randomly sampled transformation was applied to all frames to preserve temporal consistency.
Augmentations were first applied to the full $224\times224$ frames, using a shared random seed across all video frames. The quadrant crops were then extracted from the augmented frames, ensuring that all facial regions receive spatially consistent transformations.
Regularization included \textit{Label Smoothing}, \textit{Attention Dropout}, and \textit{Feed-forward Dropout}. Table \ref{table:augm_regul_training} summarizes the augmentation, regularization, and training configuration used across all experiments.

%%%%%%%%%%%%%%%%%%%%%%%%%%%%%%%%%%%%%%%%
\begin{table}
\caption{Augmentation, regularization, and training configuration.}
\label{table:augm_regul_training}
\centering
\begin{threeparttable}
\begin{tabular}{P{3.0cm} P{3.5cm}}
\toprule
Method/Parameter & Value \\
\midrule
\midrule
\textit{TrivialAugment}  & $p \in [0.20, 0.50]$ \\\hdashline
\textit{AugMix}          & $p \in [0.20, 0.50]$ \\\hdashline
\textit{Noise}           & $p \in [0.20, 0.50]$, $\sigma \leq 100$ \\\hdashline
\textit{Masking-1}       & $p \in [0.20, 0.80]$, 3 blocks \\\hdashline
\textit{Masking-2}       & $p \in [0.20, 0.50]$, 20 blocks \\\hdashline
\textit{Label Smoothing} & 0.10 \\\hdashline
\textit{Att-Dropout}     & 0.10 \\\hdashline
\textit{FF-Dropout}      & 0.10 \\
\midrule
\midrule
Optimizer       & \textit{AdamW} \\\hdashline
Learning rate   & \textit{2e-5} \\\hdashline
LR decay        & \textit{cosine} \\\hdashline
Weight decay    & 0.1 \\\hdashline
Epochs          & 100 \\\hdashline
Warmup epochs   & 15 \\\hdashline
Cooldown epochs & 5 \\\hdashline
Batch size      & 32 \\
\bottomrule
\end{tabular}
\begin{tablenotes}
\scriptsize
\item \textit{Att-Dropout}/\textit{FF-Dropout}: dropout probability in
attention/feed-forward sublayers \space
\textit{Masking-1/2}: Cutout on video frames, square $32\times32$;
value after \textbar\ = number of blocks \space
\textit{Notes}: $x_1$--$x_2$ means we sample $p\sim\mathcal{U}(x_1,x_2)$
per sample and apply the transform with probability $p$.
$x_1$--$x_2$\textbar$y$ adds the transform parameter
(\textit{e.g.}, noise $\sigma$, number of blocks)
\end{tablenotes}
\end{threeparttable}
\end{table}

%%%%%%%%%%%%%%%%%%%%%%%%%%%%%%%%%%%%%%%%%%%%%%%%%%%%%%%%
\section{Experimental Evaluation \& Results}

The \textit{AI4Pain} dataset \cite{ai4pain_2024,rojas_hirachan_2023} is used for all experiments. It comprises facial video recordings from $65$ participants, collected at the Human--Machine Interface Laboratory of the University of Canberra, Australia. The dataset is partitioned into $41$ subjects for training, $12$ for validation, and $12$ for testing. Pain stimuli were delivered through electrodes placed on the inner forearm and dorsal side of the right hand using a transcutaneous electrical nerve stimulation (TENS) device, with two pain intensity levels: low pain, individualized according to each participant's self-defined pain threshold, and high pain, defined according to each participant's self-defined pain tolerance. Video was recorded at a frame rate of $30$ frames/second.
All experiments follow an identical three-class classification setting (Low Pain, No Pain, High Pain). Each subject has $12$ samples in the training set. Accuracy, Macro-Average Precision, and Macro-Average F1 score are used to measure performance. We use the \textit{Average}, defined as the arithmetic mean of these three scores, as the primary criterion for comparing results across experiments. All ablation results in Tables~\ref{table:single_half}--\ref{table:multi_region} are evaluated on the validation split. The final configuration was selected according to the validation \textit{Average} and then evaluated once on the $12$ held-out test subjects using the released test labels. All models were trained only on the $41$ training subjects. No training, hyperparameter tuning, or model selection was performed on the test set.

\subsection{Single-Region Experiments}
Tables \ref{table:single_half} and \ref{table:single_quad} report results for the full-face and half-face regions and for each of the four individual quadrants, respectively. In all configurations, the temporal stride $s$ varies over $\{5, 10, 20, 30\}$, controlling the number of sampled frames and, consequently, the computational and memory costs.

The best overall average ($42.06\%$) for the full-size image was achieved with an input size of $224\times224$ and a stride of $20$, using $5.21$M parameters and a computational cost of $1.00$ GFLOPs. Interestingly, although the higher-density version with a stride of $5$ has more parameters ($18.77$M) and a higher computational cost ($2.73$ GFLOPs), it has a lower average than the previous one ($39.36\%$). This may suggest that there are no benefits to densely sampling time for inputs of this size. With regard to the half-image configurations, the top half at a stride of $10$ achieved the highest average ($42.17\%$), but the large difference between the precision ($56.71\%$) and F1 ($28.63\%$) scores indicates some degree of class imbalance. Such a bias would be penalized by macro-averaging for each configuration and, accordingly, it still held first place under that measure. The bottom half reached its highest average ($42.66\%$) at a stride of $20$ with $2.95$M parameters and a computational cost of $0.54$ GFLOPs. More importantly, both half-face configurations achieved performance equivalent to or better than the full-face configuration while incurring a significantly reduced computational cost, indicating substantial advantages of dividing the input into spatially separated sub-regions, even at the lowest possible resolution.

Regarding the individual quadrant regions, the top left region had the highest average score ($41.30\%$) at the largest density stride (stride $= 5$), followed by the bottom left region (average score $= 40.78\%$), then the top right region (average score $= 40.24\%$), and lastly the bottom right region (lowest average score $= 38.03\%$). As in the top-half case, precision and accuracy did not always reach their highest scores at the same stride. Although all four single-quadrant versions had a very small computational cost compared to the others ($<0.75$ GFLOPs and $<5.23$M parameters at the most dense stride), they also had among the smallest GPU latencies ($<3.66$ ms), so they should be especially suitable for applications where resources are limited. Additionally, the optimal stride differed among the four quadrants: the top-left region performed best with a stride of $5$, while the top-right region peaked at $30$. These differences may reflect the diverse expressivity and intensity of the corresponding facial regions.

\subsection{Multi-Region Experiments}

Table \ref{table:multi_region} reports results for combining multiple regions via channel concatenation, evaluating two configurations: the vertical split (Top$|$Bottom) and the full four-quadrant split (TL$|$TR$|$BL$|$BR).
The Top$|$Bottom fusion achieves a best average of $43.19\%$ at stride $20$ with $5.22$M parameters and $0.83$ GFLOPs. 
There is a clear improvement in accuracy ($44.03\%$), precision ($44.73\%$), and F1 ($40.81\%$) across all individual half-faces when using these settings. Furthermore, each metric shows a simultaneous increase. At stride $30$, we observe that the accuracy reaches $44.12\%$, while the average is $42.96\%$ using only $0.64$ GFLOPs. This would be a good compromise between cost and performance, especially if there are computational or speed constraints.

\begin{figure}
\begin{center}
\includegraphics[scale=0.47]{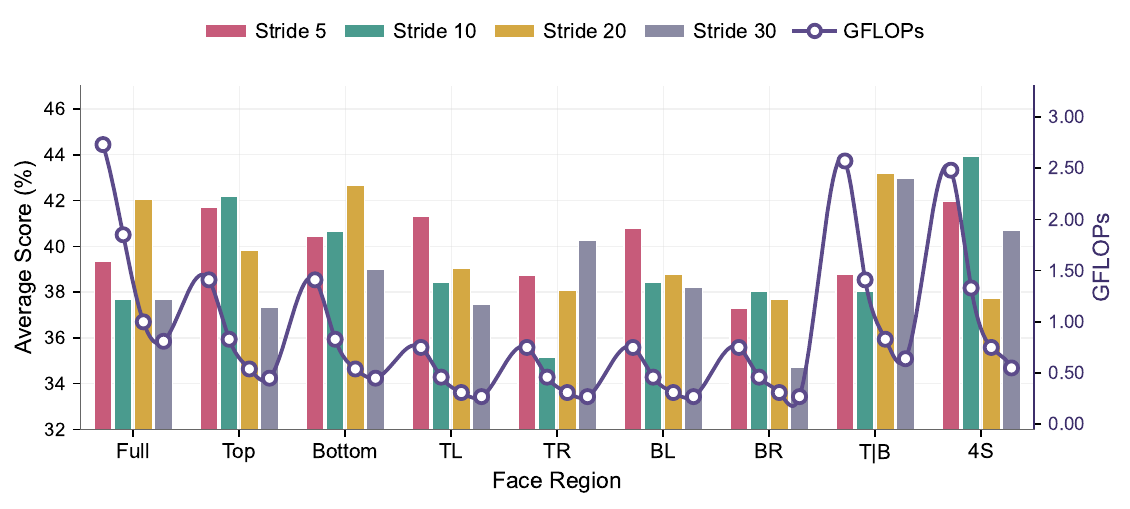}
\end{center}
\caption{Average score (\%) and GFLOPs across all face region configurations
and temporal strides. The four-quadrant fusion (4S) achieves the highest
overall performance, while single quadrants operate at the lowest cost.}
\label{comparison}
\end{figure}

The four-quadrant fusion (TL$|$TR$|$BL$|$BR) produced the best average across all tested configurations, at $43.94\%$. This was achieved with a stride of $10$, corresponding to $9.77$M parameters and $1.33$ GFLOPs. Using this configuration resulted in an accuracy of $45.00\%$, a precision of $44.54\%$, and an F1 of $42.29\%$, with all three metrics improving. However, it is also important to note that the advantages of this configuration are not universal across settings. For example, when the stride is increased to $20$, the average results across the four quadrants ($37.74\%$) were lower than those for a full-face evaluation ($42.06\%$).
Increasing temporal density to a stride of $5$ or reducing it to $30$ also underperforms, suggesting that the four-quadrant setting benefits from a specific temporal sampling range.
Figure~\ref{comparison} shows the average score for each configuration and stride, along with the GFLOPs for each. The four-quadrant fusion (4S) has the highest average scores, with individual quadrants being the least computationally expensive. As expected, GFLOPs are also very similar to the number of sampled frames and, as such, represent the most significant contributor to computational cost with respect to temporal stride.

%%%%%%%%%%%%%%%%%%%%%%%%%%%%%%%%%%%%%%%%%%%%%%%%%%%%%%%%
\begin{table*}
\caption{Performance and computational cost for the full face and half-face 
region configurations across temporal strides.}
\label{table:single_half}

\begin{center}
\begin{threeparttable}
\begin{tabular}{P{1.0cm} P{1.3cm} P{1.1cm} P{1.4cm} P{1.1cm} P{2.4cm} P{2.0cm} P{0.9cm} P{0.9cm} P{0.4cm} P{1.0cm}}
\toprule

\multicolumn{3}{c}{Input} &
\multicolumn{2}{c}{Computational Cost} &
\multicolumn{2}{c}{Inference Cost} &
\multicolumn{4}{c}{Performance} \\

\cmidrule(lr){1-3}\cmidrule(lr){4-5}\cmidrule(lr){6-7}\cmidrule(lr){8-11}
Region &Resolution &Stride & Params (M) & GFLOPs &Latency (ms) GPU$\downarrow$  &Samples/s GPU$\uparrow$ & Accuracy & Precision & F1 &\textit{Average} \\

\midrule
\midrule

Full &$224\times224$ &5  &18.77 &2.73 &12.50 &80.03  &42.76 &41.57 &33.74 &\textit{\underline{39.36}}\\\hdashline
Full &$224\times224$ &10 &9.73  &1.85 &6.08  &164.40 &38.82 &45.09 &29.10 &\textit{37.67}\\\hdashline
Full &$224\times224$ &20 &5.21  &1.00 &3.57  &279.81 &43.22 &43.68 &39.27 &\textit{\textbf{42.06}}\\\hdashline
Full &$224\times224$ &30 &3.70  &0.81 &2.32  &431.94 &39.01 &37.12 &36.85 &\textit{37.66}\\\midrule

Top &$112\times224$ &5  &9.74 &1.41 &6.29 &159.01 &42.16 &42.32 &40.58 &\textit{\underline{41.69}}\\\hdashline
Top &$112\times224$ &10 &5.22 &0.83 &3.23 &309.25 &41.19 &56.71 &28.63 &\textit{\textbf{42.17}}\\\hdashline
Top &$112\times224$ &20 &2.95 &0.54 &1.63 &614.45 &41.79 &42.40 &35.30 &\textit{39.83}\\\hdashline
Top &$112\times224$ &30 &2.20 &0.45 &1.22 &821.43 &37.94 &38.50 &35.57 &\textit{37.34}\\\midrule

Bottom &$112\times224$ &5  &9.74 &1.41 &6.29 &159.01 &40.95 &40.26 &40.05 &\textit{40.42}\\\hdashline
Bottom &$112\times224$ &10 &5.22 &0.83 &3.23 &309.25 &41.32 &43.21 &39.06 &\textit{\underline{40.66}}\\\hdashline
Bottom &$112\times224$ &20 &2.95 &0.54 &1.63 &614.45 &43.73 &44.32 &39.94 &\textit{\textbf{42.66}}\\\hdashline
Bottom &$112\times224$ &30 &2.20 &0.45 &1.22 &821.43 &40.32 &49.98 &26.99 &\textit{38.98}\\

\bottomrule
\end{tabular}

\begin{tablenotes}[para,flushleft]
\scriptsize
\textit{Average}: arithmetic mean of Accuracy, Precision, and F1. \space Bold values indicate the highest score within each region group; underlined values indicate the second-highest. 
The number of parameters and GFLOPs vary across stride configurations because the temporal stride $s$ determines 
$C = 3TK$, which directly affects the input projection dimension of the cross-attention layers (see Section \ref{sec:architecture}).
\end{tablenotes}
\end{threeparttable}
\end{center}
\end{table*}
%%%%%%%%%%%%%%%%%%%%%%%%%%%%%%%%%%%%%%%%%%%%%%%%%%%%%%%

%%%%%%%%%%%%%%%%%%%%%%%%%%%%%%%%%%%%%%%%%%%%%%%%%%%%%%%%
\begin{table*}
\caption{Performance and computational cost for individual quadrant region 
configurations across temporal strides.}
\label{table:single_quad}

\begin{center}
\begin{threeparttable}
\begin{tabular}{P{1.5cm} P{1.2cm} P{1.0cm} P{1.4cm} P{1.1cm} P{2.4cm} P{2.0cm} P{0.9cm} P{0.9cm} P{0.4cm} P{1.0cm}}
\toprule

\multicolumn{3}{c}{Input} &
\multicolumn{2}{c}{Computational Cost} &
\multicolumn{2}{c}{Inference Cost} &
\multicolumn{4}{c}{Performance} \\

\cmidrule(lr){1-3}\cmidrule(lr){4-5}\cmidrule(lr){6-7}\cmidrule(lr){8-11}
Region &Resolution &Stride & Params (M) & GFLOPs &Latency (ms) GPU$\downarrow$  &Samples/s GPU$\uparrow$ & Accuracy & Precision & F1 &\textit{Average} \\

\midrule
\midrule

Top-left &$112\times112$ &5   &5.23 &0.75 &3.66 &273.28  &42.48 &41.40 &40.01 &\textit{\textbf{41.30}}\\\hdashline
Top-left &$112\times112$ &10  &2.96 &0.46 &1.46 &684.00  &39.47 &38.70 &34.08 &\textit{38.42}\\\hdashline
Top-left &$112\times112$ &20  &1.82 &0.31 &1.30 &771.76  &40.91 &45.02 &31.16 &\textit{\underline{39.03}}\\\hdashline
Top-left &$112\times112$ &30  &1.44 &0.27 &1.19 &837.01  &38.50 &37.61 &36.34 &\textit{37.48}\\\midrule

Top-right &$112\times112$ &5   &5.23 &0.75 &3.66 &273.28  &41.65 &40.67 &33.80 &\textit{\underline{38.71}}\\\hdashline
Top-right &$112\times112$ &10  &2.96 &0.46 &1.46 &684.00  &40.07 &38.73 &26.80 &\textit{35.13}\\\hdashline
Top-right &$112\times112$ &20  &1.82 &0.31 &1.30 &771.76  &39.47 &38.50 &36.31 &\textit{38.09}\\\hdashline
Top-right &$112\times112$ &30  &1.44 &0.27 &1.19 &837.01  &41.46 &40.02 &39.24 &\textit{\textbf{40.24}}\\\midrule

Bottom-left &$112\times112$ &5   &5.23 &0.75 &3.66 &273.28  &40.62 &43.46 &38.27 &\textit{\textbf{40.78}}\\\hdashline
Bottom-left &$112\times112$ &10  &2.96 &0.46 &1.46 &684.00  &38.45 &39.25 &37.61 &\textit{38.44}\\\hdashline
Bottom-left &$112\times112$ &20  &1.82 &0.31 &1.30 &771.76  &40.68 &41.96 &33.73 &\textit{\underline{38.79}}\\\hdashline
Bottom-left &$112\times112$ &30  &1.44 &0.27 &1.19 &837.01  &40.62 &39.30 &34.71 &\textit{38.21}\\\midrule

Bottom-right &$112\times112$ &5   &5.23 &0.75 &3.66 &273.28  &40.54 &38.26 &33.11 &\textit{37.30}\\\hdashline
Bottom-right &$112\times112$ &10  &2.96 &0.46 &1.46 &684.00  &39.70 &42.20 &31.66 &\textit{\textbf{38.03}}\\\hdashline
Bottom-right &$112\times112$ &20  &1.82 &0.31 &1.30 &771.76  &39.56 &42.57 &30.93 &\textit{\underline{37.69}}\\\hdashline
Bottom-right &$112\times112$ &30  &1.44 &0.27 &1.19 &837.01  &36.74 &39.02 &28.71 &\textit{34.71}\\

\bottomrule
\end{tabular}
\end{threeparttable}
\end{center}
\end{table*}
%%%%%%%%%%%%%%%%%%%%%%%%%%%%%%%%%%%%%%%%%%%%%%%%%%%%%%%

%%%%%%%%%%%%%%%%%%%%%%%%%%%%%%%%%%%%%%%%%%%%%%%%%%%%%%%%
\begin{table*}
\caption{Performance and computational cost for multi-region fusion 
configurations across temporal strides.}
\label{table:multi_region}

\begin{center}
\begin{threeparttable}
\begin{tabular}{P{1.5cm} P{1.8cm} P{0.7cm} P{1.4cm} P{1.1cm} P{2.4cm} P{2.0cm} P{0.9cm} P{0.9cm} P{0.4cm} P{1.0cm}}
\toprule

\multicolumn{3}{c}{Input} &
\multicolumn{2}{c}{Computational Cost} &
\multicolumn{2}{c}{Inference Cost} &
\multicolumn{4}{c}{Performance} \\

\cmidrule(lr){1-3}\cmidrule(lr){4-5}\cmidrule(lr){6-7}\cmidrule(lr){8-11}
Region &Resolution &Stride & Params (M) & GFLOPs &Latency (ms) GPU$\downarrow$  &Samples/s GPU$\uparrow$ & Accuracy & Precision & F1 &\textit{Average} \\

\midrule
\midrule
Top\textbar Bottom &$(112\times224)^{\times2}$ &5   &18.80 &2.57 &12.24 &81.71  &39.58 &45.25 &31.48 &\textit{38.77}\\\hdashline
Top\textbar Bottom &$(112\times224)^{\times2}$ &10  &9.74  &1.41 &6.24  &160.37 &41.81 &40.90 &31.33 &\textit{38.01}\\\hdashline
Top\textbar Bottom &$(112\times224)^{\times2}$ &20  &5.22  &0.83 &3.34  &299.00 &44.03 &44.73 &40.81 &\textit{\textbf{43.19}}\\\hdashline
Top\textbar Bottom &$(112\times224)^{\times2}$ &30  &3.71  &0.64 &2.29  &436.99 &44.12 &43.23 &41.52 &\textit{\underline{42.96}}\\\midrule

TL\textbar TR\textbar BL\textbar BR &$(112\times112)^{\times4}$ &5   &18.84 &2.48 &12.19 &82.01  &43.56 &42.83 &39.45 &\textit{\underline{41.95}}\\\hdashline
TL\textbar TR\textbar BL\textbar BR &$(112\times112)^{\times4}$ &10  &9.77 &1.33 &6.06 &165.08  &45.00 &44.54 &42.29 &\textit{\textbf{43.94}}\\\hdashline
TL\textbar TR\textbar BL\textbar BR &$(112\times112)^{\times4}$ &20  &5.23 &0.75 &3.17 &315.67  &40.60 &39.42 &33.22 &\textit{37.74}\\\hdashline
TL\textbar TR\textbar BL\textbar BR &$(112\times112)^{\times4}$ &30  &3.71 &0.55 &1.96 &511.36  &40.60 &41.64 &39.80 &\textit{40.68}\\

\bottomrule
\end{tabular}
\begin{tablenotes}[para,flushleft]
\scriptsize
$\times2-4$: number of region crops combined via channel concatenation
\end{tablenotes}
\end{threeparttable}
\end{center}
\end{table*}
%%%%%%%%%%%%%%%%%%%%%%%%%%%%%%%%%%%%%%%%%%%%%%%%%%%%%%%

\section{Comparison with Existing Methods}
\label{comparison_testing}

To further verify the selected configuration, additional test-set checks were performed using the same protocol. The four-quadrant configuration achieved $56.00\%$ accuracy, outperforming the full-face baseline at the same stride ($49.67\%$) and the best full-face validation setting evaluated on the test set ($54.00\%$). The bottom-left quadrant also remained competitive, achieving $55.67\%$ accuracy with only one quarter of the pixel budget.
Table~\ref{table:ai4pain_test} presents the performance of the proposed method compared with previous techniques evaluated on the AI4Pain test data. Since previous benchmark studies report test accuracy as the common metric, this comparison is based solely on accuracy. The SVM baseline from \cite{ai4pain_2024} achieved $43.30\%$ accuracy with fNIRS only, $40.10\%$ with video only, and $41.70\%$ with both modalities. The transformer-based approach in \cite{gkikas_tsiknakis_painvit_2024} achieved $ 46.67\%$ accuracy when jointly using video and fNIRS. The 2D CNN of \cite{prajod_schiller_2024} achieved $49.00\%$, while the CNN--Transformer of \cite{vianto_2025} achieved $51.33\%$. The ensemble classifier of \cite{khan_aziz_2025} achieved $53.66\%$ using fNIRS features alone. The video-only transformer of \cite{nguyen_yang_2024} achieved $55.00\%$, and the multimodal framework of \cite{gkikas_rojas_painformer_2025} achieved $55.69\%$ by combining video and fNIRS.
The proposed approach achieved $ 56.00\%$ accuracy on the test set using video-only data with the four-quadrant fusion configuration (TL$|$TR$|$BL$|$BR) at a stride of $10$. This is the highest reported test accuracy among the methods compared under the fixed AI4Pain benchmark protocol, achieved using a single modality and without physiological sensing.

%%%%%%%%%%%%%%%%%%%%%%%%%%%%%%%%%%%%%%%%%%%%%%%%
\begin{table}
\footnotesize
\caption{Comparison of studies on the \textit{AI4Pain} testing set.}
\label{table:ai4pain_test}
\begin{center}
\begin{threeparttable}
\begin{tabular}{P{0.6cm} P{0.7cm} P{0.7cm} P{1.2cm} P{2.1cm} P{0.9cm}}
\toprule
\multirow{2}{*}{\shortstack{Study}} &
\multicolumn{2}{c}{Modality} &
\multicolumn{2}{c}{Method} &
\multirow{2}{*}{\shortstack{Accuracy}} \\
\cmidrule(lr){2-3}\cmidrule(lr){4-5}
& Video & fNIRS &Features &Model & \\
\midrule
\midrule

& \xmark     & \checkmark &   &     &43.30\\
& \checkmark & \xmark        &   &     &40.10\\
\multirow{-3}{*}{\cite{ai4pain_2024}}
& \checkmark & \checkmark &\multirow{-3}{*}{Handcrafted} & 
\multirow{-3}{*}{SVM} &41.70\\\hdashline
%%%%%%%%%%%%%%%%%%%%%%%%%%%%%%%%%%%%%%%%%%%%%%%%%%%%%%%
\cite{gkikas_tsiknakis_painvit_2024} & \checkmark & \checkmark  &Deep &Transformer      &46.67\\ \hdashline
\cite{prajod_schiller_2024}          & \checkmark & \xmark         &Deep & 2D CNN          &49.00\\ \hdashline
\cite{vianto_2025}                   & \checkmark & \checkmark  &Deep & CNN-Transformer &51.33\\ \hdashline
\cite{khan_aziz_2025}                & \xmark        & \checkmark  &Handcrafted & ENS      &53.66\\ \hdashline
\cite{nguyen_yang_2024}              & \checkmark & \xmark         &Deep & Transformer     &55.00\\ \hdashline

%%%%%%%%%%%%%%%%%%%%%%%%%%%%%%%%%%%%%%%%%%%%%%%%%%%%%%%
& \xmark        & \checkmark &    &    & 52.60\\
& \checkmark & \xmark        &    &    & 53.67\\
\multirow{-3}{*}{\cite{gkikas_rojas_painformer_2025}}
& \checkmark & \checkmark &\multirow{-3}{*}{Deep} & 
\multirow{-3}{*}{Transformer} &55.69\\\midrule
%%%%%%%%%%%%%%%%%%%%%%%%%%%%%%%%%%%%%%%%%%%%%%%%%%%%%%%

\rowcolor{mygray}
Our & \checkmark        & \xmark &   Deep  & Transformer  & 56.00$^\dagger$ \\

\bottomrule
\end{tabular}
\begin{tablenotes}
\scriptsize
\item \checkmark modality is used \space \xmark\space modality is not used \space ENS: Ensemble Classifier 
\item $\dagger$: Four-quadrant fusion (TL$|$TR$|$BL$|$BR), stride $10$
\end{tablenotes}
\end{threeparttable}
\end{center}
\end{table}
%%%%%%%%%%%%%%%%%%%%%%%%%%%%%%%%%%%%%%%%%%%%%%%%

\section{Conclusion}
This paper presented \textit{ReFace}, a video-based pipeline for pain assessment that reorganizes facial video into spatial quadrants before tokenization. The results show that the four-quadrant configuration improves performance over the full-face baseline under the proposed tokenization design, achieving $56.00\%$ accuracy on the \textit{AI4Pain} test set using video data alone. Individual quadrants also remained competitive at substantially lower computational cost, indicating potential value for resource-constrained applications.
The study is limited to a single dataset with small subject-level validation and test splits, and repeated-run statistical significance analysis was not performed. Future work should evaluate \textit{ReFace} across larger datasets, more diverse populations, and different pain-induction or clinical settings, while also examining the effect of region ordering and alternative anatomically motivated facial partitions.

\section*{Ethical Impact Statement}
All experiments are based on the \textit{AI4Pain} dataset, provided by the challenge organizers. Prior to data collection, participants had no known histories of neurological or psychiatric disorders, unstable medical conditions, chronic pain, or regular medication use. Participants were given full details of the experimental procedure, and written informed consent was required. Approval for the study protocol was received by the Human Ethics Committee of the University of Canberra (\textit{approval number: 11837}).
This study and the proposed framework aim to assist in the continuous objective monitoring of pain and to reduce reliance on subjective clinical judgment. However, the dataset represents a controlled experimental setting with a specific pain-induction protocol. Since demographic annotations are not officially available, subgroup fairness analysis could not be performed. Model performance may vary across demographic, facial-appearance, and clinical factors. Therefore, further validation on larger, demographically annotated, and clinically diverse datasets is required before any real-world clinical deployment. The facial image shown throughout this document is an illustrative synthetic image and does not represent any actual individual.

\section*{Acknowledgments}
The authors used large language model (LLM)-based tools for language editing and improvement. All scientific content, results, and conclusions are solely the work of the authors.

\bibliographystyle{IEEEtran}
\bibliography{library}

@article{gkikas_tsiknakis_slr_2023,
title = {Automatic assessment of pain based on deep learning methods: A systematic review},
journal = {Computer Methods and Programs in Biomedicine},
volume = {231},
pages = {107365},
year = {2023},
issn = {0169-2607},
doi = {https://doi.org/10.1016/j.cmpb.2023.107365},
author = {Stefanos Gkikas and Manolis Tsiknakis},
}

@inproceedings{gkikas_chatzaki_2022,
author = {Stefanos Gkikas and Chariklia Chatzaki and Elisavet Pavlidou and Foteini Verigou and Kyriakos Kalkanis and Manolis Tsiknakis},
doi = {10.5220/0010971700003188},
isbn = {978-989-758-566-1},
institution = {INSTICC},
journal = {Proceedings of the 8th International Conference on Information and Communication Technologies for Ageing Well and e-Health - ICT4AWE,},
pages = {155-162},
publisher = {SciTePress},
title = {Automatic Pain Intensity Estimation based on Electrocardiogram and Demographic Factors},
year = {2022},
}

@inproceedings{gkikas_chatzaki_2023,
author = {Gkikas, Stefanos and Chatzaki, Chariklia and Tsiknakis, Manolis},
title = {Multi-task Neural Networks for Pain Intensity Estimation Using Electrocardiogram and Demographic Factors},
booktitle = {Information and Communication Technologies for Ageing Well and e-Health},
year = {2023},
publisher = {Springer Nature Switzerland},
pages = {324--337},
isbn = {978-3-031-37496-8},
doi = {10.1007/978-3-031-37496-8_17},
}

@inproceedings{gkikas_tsiknakis_embc,
author={Gkikas, Stefanos and Tsiknakis, Manolis},
booktitle={2023 45th Annual International Conference of the IEEE Engineering in Medicine \& Biology Society (EMBC)}, 
title={A Full Transformer-based Framework for Automatic Pain Estimation using Videos}, 
year={2023},
pages={1-6},
doi={10.1109/EMBC40787.2023.10340872}
}

@article{gkikas_tachos_2024,
author={Gkikas, Stefanos and Tachos, Nikolaos S. and Andreadis, Stelios and Pezoulas, Vasileios C. and Zaridis, Dimitrios and Gkois, George and Matonaki, Anastasia and Stavropoulos, Thanos G. and Fotiadis, Dimitrios I.},    
title={Multimodal automatic assessment of acute pain through facial videos and heart rate signals utilizing transformer-based architectures},      
journal={Frontiers in Pain Research},      
volume={5},           
year={2024},      
DOI={10.3389/fpain.2024.1372814},      
ISSN={2673-561X}
}

@INPROCEEDINGS{gkikas_tsiknakis_thermal_2024,
author={Gkikas, Stefanos and Tsiknakis, Manolis},
booktitle={2024 12th International Conference on Affective Computing and Intelligent Interaction Workshops and Demos (ACIIW)}, 
title={Synthetic Thermal and RGB Videos for Automatic Pain Assessment Utilizing a Vision-MLP Architecture}, 
year={2024},
pages={4-12},
doi={10.1109/ACIIW63320.2024.00006}
}

@INPROCEEDINGS{gkikas_tsiknakis_painvit_2024,
author={Gkikas, Stefanos and Tsiknakis, Manolis},
booktitle={2024 12th International Conference on Affective Computing and Intelligent Interaction Workshops and Demos (ACIIW)}, 
title={Twins-PainViT: Towards a Modality-Agnostic Vision Transformer Framework for Multimodal Automatic Pain Assessment Using Facial Videos and fNIRS}, 
year={2024},
pages={13-21},
doi={10.1109/ACIIW63320.2024.00007}
}

@misc{gkikas_phd_thesis_2025,
title={A Pain Assessment Framework based on multimodal data and Deep Machine Learning methods}, 
author={Stefanos Gkikas},
year={2025},
note = {arXiv preprint arXiv:2505.05396},
eprint={2505.05396},
archivePrefix={arXiv},
primaryClass={cs.AI},
url={https://arxiv.org/abs/2505.05396}
}

@article{meehan_mcrae_1995,
title={Analgesic administration, pain intensity, and patient satisfaction in cardiac surgical patients},
author={Meehan, DA and McRae, ME and Rourke, DA and Eisenring, C and Imperial, FA},
journal={American Journal of Critical Care},
volume={4},
number={6},
pages={435--442},
year={1995},
publisher={American Association of Critical Care Nurses}
}

@article{puntilo_staannard_2022,
title = {Use of a pain assessment and intervention notation (P.A.I.N.) tool in critical care nursing practice: Nurses' evaluations},
journal = {Heart \& Lung},
volume = {31},
number = {4},
pages = {303-314},
year = {2002},
issn = {0147-9563},
doi = {https://doi.org/10.1067/mhl.2002.125652},
author = {Kathleen A. Puntillo and Daphne Stannard and Christine Miaskowski and Karen Kehrle and Sheila Gleeson},
}

@article{breivik_eisenberg_2013,
title={The individual and societal burden of chronic pain in Europe: the case for strategic prioritisation and action to improve knowledge and availability of appropriate care},
author={Breivik, Harald and Eisenberg, Elon and O’Brien, Tony},
journal={BMC public health},
volume={13},
pages={1--14},
year={2013},
publisher={Springer},
doi={https://doi.org/10.1186/1471-2458-13-1229}
}

@article{rojas_brown_2023,
title={A systematic review of neurophysiological sensing for the assessment of acute pain},
author={Fernandez Rojas, Raul and Brown, Nicholas and Waddington, Gordon and Goecke, Roland},
journal={NPJ Digital Medicine},
volume={6},
number={1},
pages={76},
doi = {https://doi.org/10.1038/s41746-023-00810-1},
year={2023},
publisher={Nature Publishing Group UK London}
}

@article{khan_umar_2025,
author = {Khan, Muhammad Umar and Chetty, Girija and Goecke, Roland and Fernandez-Rojas, Raul},
title = {A Systematic Review of Multimodal Signal Fusion for Acute Pain Assessment Systems},
year = {2025},
publisher = {Association for Computing Machinery},
address = {New York, NY, USA},
issn = {0360-0300},
doi = {10.1145/3737281},
journal = {ACM Comput. Surv.},
}

@ARTICLE{gkikas_rojas_painformer_2025,
author={Gkikas, Stefanos and Rojas, Raul Fernandez and Tsiknakis, Manolis},
journal={IEEE Transactions on Affective Computing}, 
title={PainFormer: A Vision Foundation Model for Automatic Pain Assessment}, 
year={2025},
volume={16},
number={4},
pages={3369-3386},
doi={10.1109/TAFFC.2025.3605475}
}

@article{rojas_hirachan_2023,
author={Fernandez Rojas, Raul and Hirachan, Niraj and Brown, Nicholas and Waddington, Gordon and Murtagh, Luke and Seymour, Ben and Goecke, Roland},    
title={Multimodal physiological sensing for the assessment of acute pain},      
journal={Frontiers in Pain Research},      
volume={4},           
year={2023},              
doi={10.3389/fpain.2023.1150264},      
issn={2673-561X}
}

@article{huang_dong_2022,
author={Huang, Dong and Feng, Xiaoyi and Zhang, Haixi and Yu, Zitong and Peng, Jinye and Zhao, Guoying and Xia, Zhaoqiang},
journal={IEEE Transactions on Multimedia}, 
title={Spatio-Temporal Pain Estimation Network With Measuring Pseudo Heart Rate Gain}, 
year={2022},
volume={24},
number={},
pages={3300-3313},
doi={10.1109/TMM.2021.3096080}
}

@inproceedings{ai4pain_2024,
author={Fernandez–Rojas, Raul and Joseph, Calvin and Hirachan, Niraj and Seymour, Ben and Goecke, Roland},
booktitle={2024 12th International Conference on Affective Computing and Intelligent Interaction Workshops and Demos (ACIIW)}, 
title={The AI4Pain Grand Challenge 2024: Advancing Pain Assessment with Multimodal fNIRS and Facial Video Analysis}, 
year={2024},
pages={55-60},
doi={10.1109/ACIIW63320.2024.00012}
}

@INPROCEEDINGS{prajod_schiller_2024,
author={Prajod, Pooja and Schiller, Dominik and Don, Daksitha Withanage and André, Elisabeth},
booktitle={2024 12th International Conference on Affective Computing and Intelligent Interaction Workshops and Demos (ACIIW)}, 
title={Faces of Experimental Pain: Transferability of Deep-Learned Heat Pain Features to Electrical Pain*}, 
year={2024},
volume={},
number={},
pages={31-38},
doi={10.1109/ACIIW63320.2024.00009}
}

@misc{nguyen_yang_2024,
title={Transformer with Leveraged Masked Autoencoder for video-based Pain Assessment}, 
author={Minh-Duc Nguyen and Hyung-Jeong Yang and Soo-Hyung Kim and Ji-Eun Shin and Seung-Won Kim},
year={2024},
eprint={2409.05088},
archivePrefix={arXiv},
primaryClass={cs.CV},
}

@article{khan_aziz_2025,
title = {Empirically Transformed Energy Patterns: A novel approach for capturing fNIRS signal dynamics in pain assessment},
journal = {Computers in Biology and Medicine},
volume = {192},
pages = {110300},
year = {2025},
issn = {0010-4825},
doi = {https://doi.org/10.1016/j.compbiomed.2025.110300},
author = {Muhammad Umar Khan and Sumair Aziz and Luke Murtagh and Girija Chetty and Roland Goecke and Raul {Fernandez Rojas}},
}

@Article{vianto_2025,
AUTHOR = {Vianto, Jo and Divakaran, Anjitha and Yang, Hyungjeong and Yeom, Soonja and Kim, Seungwon and Kim, Soohyung and Shin, Jieun},
TITLE = {Multimodal Model for Automated Pain Assessment: Leveraging Video and fNIRS},
JOURNAL = {Applied Sciences},
VOLUME = {15},
YEAR = {2025},
NUMBER = {9},
ISSN = {2076-3417},
DOI = {10.3390/app15095151}
}

@article{kaye_jones_2017,
title={Prescription opioid abuse in chronic pain: an updated review of opioid abuse predictors and strategies to curb opioid abuse: part 1},
author={Kaye, Alan David and Jones, Mark R and Kaye, Adam M and Ripoll, Juan G and Galan, Vincent and Beakley, Burton D and Calixto, Francisco and Bolden, Jamie L and Urman, Richard D and Manchikanti, Laxmaiah},
journal={Pain physician},
volume={20},
number={2},
pages={S93},
year={2017}
}

@article{stampas_pedroza_2020,
title={The first 24 h: opioid administration in people with spinal cord injury and neurologic recovery},
author={Stampas, Argyrios and Pedroza, Claudia and Bush, Jennifer N and Ferguson, Adam R and Kramer, John L Kipling and Hook, Michelle},
journal={Spinal Cord},
volume={58},
number={10},
pages={1080--1089},
year={2020},
publisher={Nature Publishing Group UK London}
}

@article{benyamin_trescot_2008,
title={Opioid complications and side effects},
author={Benyamin, Ramsin and Trescot, Andrea M and Datta, Sukdeb and Buenaventura, Ricardo M and Adlaka, Rajive and Sehgal, Nalini and Glaser, Scott E and Vallejo, Ricardo},
journal={Pain physician},
volume={11},
number={2S},
pages={S105},
year={2008},
publisher={American Society of Interventional Pain Physician}
}

@inproceedings{gkikas_kyprakis_eda_2025,
author = {Gkikas, Stefanos and Kyprakis, Ioannis and Tsiknakis, Manolis},
title = {Multi-Representation Diagrams for Pain Recognition: Integrating Various Electrodermal Activity Signals into a Single Image},
year = {2025},
isbn = {9798400720765},
publisher = {Association for Computing Machinery},
address = {New York, NY, USA},
doi = {10.1145/3747327.3764793},
booktitle = {Companion Proceedings of the 27th International Conference on Multimodal Interaction},
pages = {162–171},
numpages = {10},
series = {ICMI Companion '25}
}

@inproceedings{gkikas_kyprakis_resp_2025,
author = {Gkikas, Stefanos and Kyprakis, Ioannis and Tsiknakis, Manolis},
title = {Efficient Pain Recognition via Respiration Signals: A Single Cross-Attention Transformer Multi-Window Fusion Pipeline},
year = {2025},
isbn = {9798400720765},
publisher = {Association for Computing Machinery},
address = {New York, NY, USA},
doi = {10.1145/3747327.3764782},
booktitle = {Companion Proceedings of the 27th International Conference on Multimodal Interaction},
pages = {70–79},
numpages = {10},
series = {ICMI Companion '25}
}

@inproceedings{gkikas_tiny_2025,
author = {Gkikas, Stefanos and Kyprakis, Ioannis and Tsiknakis, Manolis},
title = {Tiny-BioMoE: a Lightweight Embedding Model for Biosignal Analysis},
year = {2025},
isbn = {9798400720765},
publisher = {Association for Computing Machinery},
address = {New York, NY, USA},
doi = {10.1145/3747327.3764788},
booktitle = {Companion Proceedings of the 27th International Conference on Multimodal Interaction},
pages = {117–126},
numpages = {10},
series = {ICMI Companion '25}
}

@article{bargshady_aziz_2025,
author = {Bargshady, Ghazal and Aziz, Sumair and Gkikas, Stefanos and Tsiknakis, Manolis and Goecke, Roland and Fernandez Rojas, Raul},
title = {Pain Assessment Using Multi-Kernel-FCN-LSTM and Haemoglobin Difference in fNIRS},
year = {2025},
publisher = {Association for Computing Machinery},
address = {New York, NY, USA},
doi = {10.1145/3757931},
journal = {ACM Trans. Comput. Healthcare}
}

@article{farmani_bargshady_2025,
author    = {Farmani, Jaleh and Bargshady, Ghazal and Gkikas, Stefanos and Tsiknakis, Manolis and Fernandez Rojas, Raul},
title     = {A CrossMod-Transformer deep learning framework for multi-modal pain detection through EDA and ECG fusion},
journal   = {Scientific Reports},
year      = {2025},
volume    = {15},
number    = {1},
pages     = {29467},
doi       = {10.1038/s41598-025-14238-y},
url       = {https://doi.org/10.1038/s41598-025-14238-y},
issn      = {2045-2322},
publisher = {Springer Nature}
}

@ARTICLE{khera_rangasamy_2021, 
AUTHOR={Khera, Tanvi  and Rangasamy, Valluvan },       
TITLE={Cognition and Pain: A Review},      
JOURNAL={Frontiers in Psychology},        
VOLUME={Volume 12 - 2021},
YEAR={2021},
DOI={10.3389/fpsyg.2021.673962},
ISSN={1664-1078},
}

@inproceedings{trivialAugment,
  author={Müller, Samuel G. and Hutter, Frank},
  booktitle={2021 IEEE/CVF International Conference on Computer Vision (ICCV)}, 
  title={TrivialAugment: Tuning-free Yet State-of-the-Art Data Augmentation}, 
  year={2021},
  volume={},
  number={},
  pages={754-762},
  doi={10.1109/ICCV48922.2021.00081}
  }

@article{augmix,
  title={Augmix: A simple data processing method to improve robustness and uncertainty},
  author={Hendrycks, Dan and Mu, Norman and Cubuk, Ekin D and Zoph, Barret and Gilmer, Justin and Lakshminarayanan, Balaji},
  journal={arXiv preprint arXiv:1912.02781},
  year={2019}
}

@article{zhang_2016,
  title={Joint face detection and alignment using multitask cascaded convolutional networks},
  author={Zhang, Kaipeng and Zhang, Zhanpeng and Li, Zhifeng and Qiao, Yu},
  journal={IEEE signal processing letters},
  volume={23},
  number={10},
  pages={1499--1503},
  year={2016},
  publisher={IEEE}
}

@article{usa_bdc_2013,
author = {US Burden of Disease Collaborators},
title = "{The State of US Health, 1990-2010: Burden of Diseases, Injuries, and Risk Factors}",
journal = {JAMA},
volume = {310},
number = {6},
pages = {591-606},
year = {2013},
month = {08},
issn = {0098-7484},
doi = {10.1001/jama.2013.13805},
}

@article{gaskin_richard_2012,
title = {The Economic Costs of Pain in the United States},
journal = {The Journal of Pain},
volume = {13},
number = {8},
pages = {715-724},
year = {2012},
issn = {1526-5900},
doi = {https://doi.org/10.1016/j.jpain.2012.03.009},
author = {Darrell J. Gaskin and Patrick Richard},
}

@article{dinakar_stillman_2016,
  author = {Dinakar, Pradeep and Stillman, Alexandra Marion},
  doi = {10.1016/J.SPEN.2016.10.003},
  issn = {1071-9091},
  journal = {Seminars in Pediatric Neurology},
  month = {aug},
  number = {3},
  pages = {201--208},
  publisher = {W.B. Saunders},
  title = {Pathogenesis of Pain},
  volume = {23},
  year = {2016}
}

@article {abdulla_adams_2013,
Title = {Guidance on the management of pain in older people},
Author = {Abdulla, Aza and Adams, Nicola and Bone, Margaret and Elliott, Alison M and Gaffin, Jean and Jones, Derek and Knaggs, Roger and Martin, Denis and Sampson, Liz and Schofield, Pat and {British Geriatric Society}},
DOI = {10.1093/ageing/afs200},
Volume = {42 Suppl 1},
Month = {March},
Year = {2013},
Journal = {Age and ageing},
ISSN = {0002-0729},
Pages = {i1—57},
}

@article{tavakolian_hadid_2019,
   author = {Tavakolian, Mohammad and Hadid, Abdenour},
   doi = {10.1007/s11263-019-01191-3},
   issn = {15731405},
   journal = {International Journal of Computer Vision},
   month = {oct},
   number = {10},
   pages = {1413--1425},
   publisher = {Springer New York LLC},
   title = {A Spatiotemporal Convolutional Neural Network for Automatic Pain Intensity Estimation from Facial Dynamics},
   volume = {127},
   year = {2019}
}

@inproceedings{werner_hamadi_2014,
   author = {Philipp Werner and Ayoub Al-Hamadi and Robert Niese and Steffen Walter and Sascha Gruss and Harald C. Traue},
   doi = {10.1109/ICPR.2014.784},
   isbn = {9781479952083},
   issn = {10514651},
   journal = {International Conference on Pattern Recognition},
   pages = {4582-4587},
   publisher = {Institute of Electrical and Electronics Engineers Inc.},
   title = {Automatic pain recognition from video and biomedical signals},
   year = {2014},
}

@inproceedings{werner_hamadi_walter_2017,
  author={Werner, Philipp and Al-Hamadi, Ayoub and Walter, Steffen},
  booktitle={2017 Seventh International Conference on Affective Computing and Intelligent Interaction Workshops and Demos (ACIIW)}, 
  title={Analysis of facial expressiveness during experimentally induced heat pain}, 
  year={2017},
  pages={176-180},
  doi={10.1109/ACIIW.2017.8272610}
}

@article{huang_xia_li_2019,
  author = {Huang, Dong and Xia, Zhaoqiang and Li, Lei and Wang, Kunwei and Feng, Xiaoyi},
  doi = {10.1117/1.jei.28.4.043008},
  issn = {1017-9909},
  journal = {Journal of Electronic Imaging},
  number = {04},
  pages = {1},
  title = {Pain-awareness multistream convolutional neural network for pain estimation},
  volume = {28},
  year = {2019}
}

@article{james_abate_2018,
  doi = {10.1016/S0140-6736(18)32279-7},
  institution = {GBD 2017 Disease and Injury Incidence and Prevalence Collaborators},
  issn = {1474547X},
  journal = {The Lancet},
  author={Collaborators, GBD and others},
  month = {nov},
  number = {10159},
  pages = {1789--1858},
  pmid = {30496104},
  title = {Global, regional, and national incidence, prevalence, and years lived with disability for 354 Diseases and Injuries for 195 countries and territories, 1990-2017: A systematic analysis for the Global Burden of Disease Study 2017},
  volume = {392},
  year = {2018}
}

@misc{gkikas_arzate_eeite_pain_2026,
title={A Lightweight Transformer for Pain Recognition from Brain Activity}, 
author={Stefanos Gkikas and Christian Arzate Cruz and Yu Fang and Lu Cao and Muhammad Umar Khan and Thomas Kassiotis and Giorgos Giannakakis and Raul Fernandez Rojas and Randy Gomez},
year={2026},
eprint={2604.16491},
archivePrefix={arXiv},
primaryClass={cs.CV}
}

@article{khan_chetty_2026,
title = {GIAFormer: A Gradient-Infused Attention and Transformer for Pain Assessment with EDA-fNIRS Fusion},
journal = {Information Fusion},
volume = {131},
pages = {104173},
year = {2026},
issn = {1566-2535},
doi = {https://doi.org/10.1016/j.inffus.2026.104173},
author = {Muhammad Umar Khan and Girija Chetty and Stefanos Gkikas and Manolis Tsiknakis and Roland Goecke and Raul Fernandez-Rojas},
}

@misc{gkikas_arzate_workload_acii_2026,
title={{Towards a Unified Modality-Agnostic Multimodal Framework for Cognitive Workload Assessment}},
author={Stefanos Gkikas and Christian Arzate Cruz and Calvin Joseph and Giorgos Giannakakis and Raul Fernandez Rojas},
year={2026},
eprint={XXXX.XXXXX},
archivePrefix={arXiv},
primaryClass={cs.CV}
}

@inproceedings{gkikas_arzate_pain_icmi_2026,
  title={{A Unified Tokenization Framework for Pain Recognition using Heterogeneous 3D Modalities}},
  author={Stefanos Gkikas and Christian Arzate Cruz and Valentina Becchetti and Muhammad Umar Khan and Alessandro Giuseppi and Raul Fernandez Rojas},
  booktitle={Proceedings of the 28th ACM International Conference on Multimodal Interaction},
  year={2026},
  location={Napoli, Italy},
  publisher={Association for Computing Machinery}
}

\end{document}